# Boosted Training of Convolutional Neural Networks for Multi-Class Segmentation

Lorenz Berger[1], Eoin Hyde[1], Matt Gibb[1], Nevil Pavithran[2], Garin Kelly[2], Faiz Mumtaz[2], Sébastien Ourselin[3]

[1]Innersight Labs, [2]Royal Free Hospital, London UK, [3]Kings College London

## Abstract

Training deep neural networks on large and sparse datasets is still challenging and can require large amounts of computation and memory. In this work, we address the task of performing semantic segmentation on large volumetric data sets, such as CT scans. Our contribution is threefold: 1) We propose a boosted sampling scheme that uses a-posterior error maps, generated throughout training, to focus sampling on difficult regions, resulting in a more informative loss. This results in a significant training speed up and improves learning performance for image segmentation. 2) We propose a novel algorithm for boosting the SGD learning rate schedule by adaptively increasing and lowering the learning rate, avoiding the need for extensive hyperparameter tuning. 3) We show that our method is able to attain new state-of-the-art results on the VISCERAL Anatomy benchmark.

## 1  Introduction

This paper addresses the problem of efficiently training convolutional neural networks (CNNs) on large and imbalanced datasets. We propose a training strategy that boosts the likelihood of selecting difficult samples from the training data, to effectively speed up training and avoid over-sampling data that contains little extra information. In this work, we investigate the problem of automatic segmentation from high resolution 3D CT scans. Several deep learning techniques [1–3] have recently been proposed for 3D segmentation of medical datasets. To overcome the problem of dealing with these large datasets, commonly of dimension $512 \times 512 \times 700$, previous approaches train a CNN on a cropped region of interest which reduces the size of individual training images by around 100 fold [2]. By reducing the size of training images, they can now be fit into memory and a network can be trained effectively on the selected data. However, identifying regions of interest requires an additional pre-processing step which may not be easy in many applications. Also, training CNNs on cropped images limits the field of view of the CNN and subsequently can introduce unwanted image boundary induced effects during testing. Other applications, where training CNNs on very large images is a problem, includes the segmentation of histology datasets or the segmentation of aerial images. For example in aerial image segmentation, training a CNN to segment ships can be difficult because large portions of the image contain water which provide little information during training, resulting in slow learning. Some ideas to address this have already been proposed, for example in [3] a fixed, hand-crafted, pre-computed weight map is used to help learn small separation borders between touching cells for biomedical image segmentation. Curriculum learning [4] and derivative methods like self-paced learning [5] build on the intuition that, rather than considering all samples simultaneously, the algorithm should be presented with the training data in a meaningful order that facilitates learning. This idea has already successfully been applied to image classification [6], by ordering images from easy to hard during training. Also in [7] a selective sampling method on negative samples was used to improve the training of a CNN for hemorrhage detection in 2D fungus images.

The focus of this paper is supervised semantic segmentation where a representative training set is available with dense manual multi-class annotations and the challenge lies in efficiently learning from large 3D datasets such as full CT scans. We present a boosted sampling algorithm that prioritises sampling patches containing higher amounts of information (high training error). However this introduces a further non-linearity into the optimization process, as the amount of information within mini-batches is ever changing, further complicating the tuning of learning rate schedule hyperparameters, which can make the difference between bad and good segmentation results [1]. To resolve this we propose an algorithm for automatically boosting and decaying the SGD learning rate schedule throughout training. Experiments are presented using the VISCERAL anatomy benchmark dataset.

## 2 Methods

**isample: Boosted sampling:** The sampling algorithm prioritises sampling patches from training data that at the time of training produces large training error, and avoids over-sampling data that contains little extra information. This simple method is described in Algorithm 1, where $\mathcal{U}(0,1)$ is a random number drawn from the uniform distribution and $\boldsymbol{E}_i$ refers to the error map of the $i^{th}$ training image. Error maps can easily be calculated, either after each epoch or concurrently to the training process, as $\boldsymbol{E}_k(\boldsymbol{x}) = 1 - \mathrm{CNN}(\boldsymbol{w}, \boldsymbol{I}_k(\boldsymbol{x}))_{\boldsymbol{L}_k(\boldsymbol{x})}$, where $\mathrm{CNN}(w, \boldsymbol{I}_k(\boldsymbol{x}))_{\boldsymbol{L}_k(\boldsymbol{x})}$ is a map of the CNN predictions over the full training image $\boldsymbol{I}_k$, evaluated using the most current weights, $\boldsymbol{w}$, and outputting the probability of the true class label $\boldsymbol{L}_k(\boldsymbol{x})$, at position $\boldsymbol{x}$.

**AutoLR: Adaptive learning rate scheduling:** To overcome expensive and time consuming hyperparameter tuning of the learning rate schedule we propose the following, simple to implement, algorithm. We evolve the learning rate throughout training, by evaluating the CNN's validation performance on a small population of concurrent CNN runs, each trained with a different learning rate, for some validation period. The weights of the best performing run at the end of the period are then used to spawn a new set (population) of CNN runs, each with a different learning rate. This simple procedure is outlined in more detail in

**Algorithm 1** isample: Boosted sampling
---
Initialise error maps for every image in the training data: $\boldsymbol{E}_i(x) = 1$.
**while** CNN training **do**
    **while** training for 1 epoch **do**
        **while** filling batch with patches **do**
            Pick an image $\boldsymbol{I}_j$ from the training set $\boldsymbol{I}^*$.
            Pick a class $k$ from the corresponding label map $\boldsymbol{L}_j$.
            Pick a patch in image $\boldsymbol{I}_j$, centered at location $\boldsymbol{c}$, where $\boldsymbol{L}_j(\boldsymbol{c}) = k$.
            Accept patch into batch if $\boldsymbol{E}_i(\boldsymbol{c}) > \mathcal{U}(0,1)$.
        **end while**
        Back-propagate loss of batch and update the current CNN weights: $\boldsymbol{w}$.
    **end while**
    Select a subset of images, $\boldsymbol{I}^*$, and label maps, $\boldsymbol{L}^*$, from the training set:
    **for** $[\boldsymbol{I}_k, \boldsymbol{L}_k] \in [\boldsymbol{I}^*, \boldsymbol{L}^*]$ **do**
        Update error maps: $\boldsymbol{E}_k(\boldsymbol{x}) = 1 - \text{CNN}(\boldsymbol{w}, \boldsymbol{I}_k(\boldsymbol{x}))_{\boldsymbol{L}_k(\boldsymbol{x})}$
    **end for**
**end while**

Algorithm 2. The number of parallel runs used can be chosen arbitrarily. However a clear disadvantage, as with any genetic type algorithm, is that the more parallel runs are spawned, the more parallel processing is required. The described algorithm can easily be generalised to fit within the framework of evolutionary algorithms and therefore be extended to deal with more complex genetic operators.

**Algorithm 2** AutoLR: Adaptive learning rate scheduler
---
Choose, $\beta$ the step length, the number of runs, $R_i$, and the initial learning rate and modulating factor for each run $\eta_i, \lambda_i$.
**while** CNN training **do**
    **while** training each run, $R_i$ in parallel, for $\gamma$ epochs **do**
        Record validation performance of each run, $R_i$, at each epoch in the vector $\boldsymbol{v}_i$ of length $\beta$.
    **end while**
    Identify the best performing run, $R_i^* = \max\limits_{R_i}\{\max[\boldsymbol{v}_i]\}$
    Set the optimal learning rate to be the learning rate of the best performing run, $\eta^* = R_i^*, \eta$
    Update the learning rate of each run, $\eta_i = \eta^* \lambda_i$
**end while**

**CNN setup:** A detailed account is given in the Appendix.

## 3 Results

For illustrative purposes, the first experiment focuses on segmenting only the kidneys from full body CT scans to investigate the performance of the boosted sampling algorithm referred to as 'isample'. We then present results on multi-class organ segmentation from CT data and investigate the performance of the automated learning rate scheduler, referred to as 'AutoLR'. References for all benchmarking methods are annotated with a star and refer to the corresponding reference number in [8].

**Segmenting kidneys from full body CT scans.** We evaluate our method on contrast enhanced CT scans from the VISCERAL Anatomy 3 dataset, made up of 20 training scans, and 10 unseen testing scans (currently not available to download) [8]. For this experiment, we randomly split the training set into 16 scans for training (80%) and 4 scans for validation (20%). We also present results of our online submission on the unseen test dataset. We use labels for kidneys to train the CNN, resulting in a simple two-class, foreground (kidneys) and background (everything else), segmentation problem. The learning rate is set at 0.001, and the batch size was 20 patches. Figure 1 shows coronal slices of a training error volume $\boldsymbol{E}_k(\boldsymbol{x})$. As seen in Figure 1c, initially there is significant error produced by the CNN prediction at epoch 16, for example misclassifying the aorta (part of the background class) because of similar intensity values to the kidneys. After more training, at epoch 50, Figure 1d shows that the error is much lower. The CNN has now learned that the aorta is part of the background class. However more subtle regions such as the collecting system and large vessels within the kidney (see small hole in the true segmentation of the left kidney in Figure 1b) still produce high errors, and further focused training is required to optimize the weights until they are correctly classified. There also remains high error around the border of the kidneys, which will result in the sampling process selecting more patches from this border region. Coincidentally, this results in our method learning to train the network with a similar loss to the hand-crafted border weighted loss function designed in [3]. Because isample adaptively selects and boosts more difficult patches as training progresses, the loss is higher, as seen in Figure 2a. In Figure 2b, the sampler achieves faster generalisation, and our current results indicate that the final generalisation of the CNN trained with the proposed sampling scheme is slightly improved for this sparse segmentation setup, where the kidneys only make up $\sim 0.3\%$ of the voxels within the whole scan. Further, when using the isample scheme the CNN is able to achieve a Dice score of 0.855 after only $5k$ training iterations. This is close to the end of training performance, a Dice score of 0.899 after 40k, achieved by the Dual CNN without isample in use. Table 2c shows Dice scores for segmenting both kidneys using different methods. The proposed method with isample performs significantly better than without. We also submitted our method, with the addition of a CRF [1] as a post-processing step, to segment the test dataset, and achieved the top score for segmenting the left and right kidneys. Inference on a full size CT scans takes $\sim 65$ seconds using four Tesla K40 GPU cards, each with 4GB of RAM.

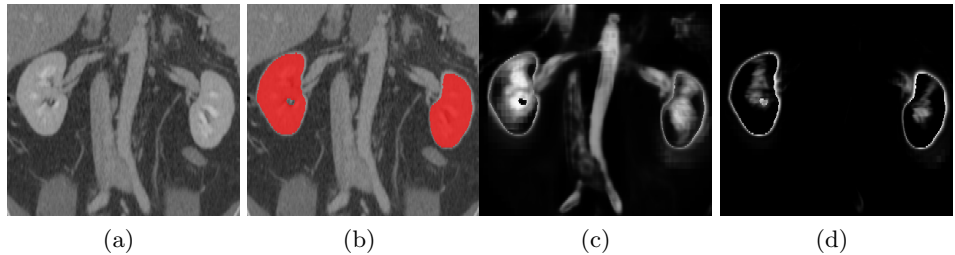

(a) (b) (c) (d)

Fig. 1: Coronal slices: (a) Raw CT scan from the training set (b) Kidney segmentation overlaid onto scan (c) Error map, $\boldsymbol{E}_k(\boldsymbol{x})$, of foreground and background classification on a training scan after 16 epochs. (d) $\boldsymbol{E}_k(\boldsymbol{x})$ after 50 epochs. For the error maps, white corresponds to voxels that are incorrectly classified and black to correctly classified voxels.

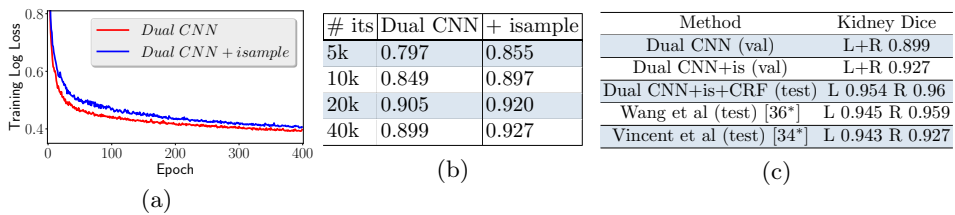

| # its | Dual CNN | + isample |
|---|---|---|
| 5k | 0.797 | 0.855 |
| 10k | 0.849 | 0.897 |
| 20k | 0.905 | 0.920 |
| 40k | 0.899 | 0.927 |

| Method | Kidney Dice |
|---|---|
| Dual CNN (val) | L+R 0.899 |
| Dual CNN+is (val) | L+R 0.927 |
| Dual CNN+is+CRF (test) | L 0.954 R 0.96 |
| Wang et al (test) [36*] | L 0.945 R 0.959 |
| Vincent et al (test) [34*] | L 0.943 R 0.927 |

(a) (b) (c)

Fig. 2: (a) Training loss averaged over 3 runs, with and without using isample. (b) Mean Dice scores, averaged over three separate runs, at different number of iterations, throughout training. (c) Dice scores for different automatic kidney segmentation methods on the VISCERAL test and validation datasets.

**Multi-class organ segmentation.** We now extend the model to include a multi-class classification output and trained on the main organs available on the VISCERAL contrast CT dataset. We randomly split the training set into 14 scans for training (70%), 2 scans for validation (10%) and 4 scans for hold-out testing (20%). Figures 3a and 3b demonstrate how AutoLR adapts to find an optimal learning rate schedule. For this example we only show results of the first 200 epochs of training to best visualise the method's dynamics. Figure 3b shows the learning rate throughout training by the three spawned runs. After training for 50 epochs, the recorded validation scores are evaluated, which for these experiments is the mean Dice score of all the segmented organs on the full validation scans. In Figure 3a it can be seen that the maximum Dice score for this first period was achieved by the run colored in red, training with a learning rate of $\eta = 0.05$, such that $\eta* = 0.05$ in Algorithm 2. Therefore after the first step period, following epoch 50, the three runs evolve such that $\eta_0 = \eta^* \lambda_0 = 0.05 \times 2 = 0.1$, $\eta_1 = 0.05 \times 1 = 0.05$ and $\eta_2 = 0.05 \times 0.5 = 0.025$ throughout the next step period. From Figure 3b we can see that this causes the learning rate to initially increase, which speeds up training and allows SGD to move away from its initialisation. This is followed by AutoLR reducing the learning rate. The loss of the three runs is shown in Figure 3c. The higher the learning rate the more stochasticity can be seen in the SGD loss function. The batch size was increased to 40 patches. To benchmark the proposed learning rate scheduling Algorithm 2 we selected several handcrafted learning rate training schedules and compare their performance against the proposed method. Before arriving at these handcrafted many initial experiments (not shown here) were run to get close to the optimal initial learning rates and decay schedules shown in Figures 3d and 3e. To produce the results shown in Figure 3, five weeks of training time using eight NVIDIA Tesla K40s was required. We find that AutoLR performs competitively with the top handcrafted learning rate schedules, shown in Figure 3d and 3e, and the automatically found learning rate schedule follows a similar path to the top performing handcrafted schedule, as shown in Figure 3f. It can be observed from Figures 3d and 3e that different learning rate schedules are preferred when using isample, than when without isample. This highlights the importance of tuning the learning rate schedule throughout development and motivates the use of an automated method. For additional accuracy gains we post-process the segmentation output by only retaining the largest connected binary object of each organ. Dice scores achieved by our method on the hold out test set, and other state-of-the-art methods are given in Table 1.

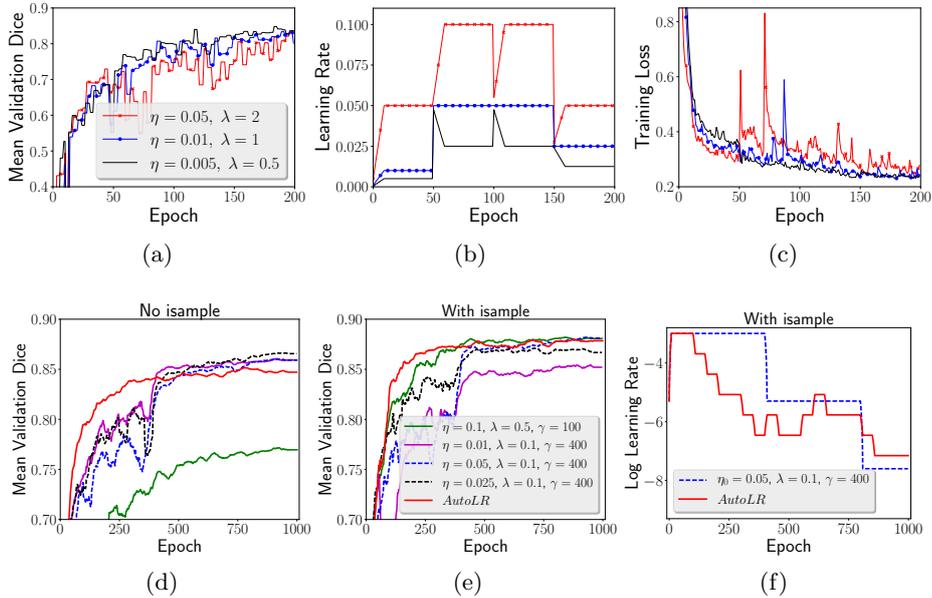

Fig. 3: Mean organ validation Dice score (a), learning rate evolution (b) and log loss (c) of three parallel runs during AutoLR. Validation performance of handcrafted and AutoLR (d,e), and learning rate path found by AutoLR (f).

|  | Aorta | Lung | Kidney | PMajor | Liver | Abdom | Spleen | Sternum | Trachea | Bladder |
|---|---|---|---|---|---|---|---|---|---|---|
| Gass et al [38*] | 0.785 | 0.963 | 0.914 | 0.813 | 0.908 | - | 0.781 | 0.635 | 0.847 | 0.683 |
| Jimenez et al [40*] | 0.762 | 0.961 | 0.899 | 0.797 | 0.887 | 0.463 | 0.730 | 0.721 | 0.855 | 0.679 |
| Kéchichian et al [43*] | 0.681 | 0.966 | 0.912 | 0.802 | 0.933 | 0.538 | 0.895 | 0.713 | 0.824 | 0.823 |
| Vincent et al [34*] | 0.838 | 0.972 | 0.935 | 0.869 | 0.942 | - | - | - | - | - |
| AutoLR (Dual CNN) | 0.834 | 0.972 | 0.913 | 0.815 | 0.914 | 0.804 | 0.778 | 0.765 | 0.828 | 0.780 |
| AutoLR + isample | 0.875 | 0.979 | 0.922 | 0.851 | 0.929 | 0.824 | 0.895 | 0.829 | 0.910 | 0.860 |
| AutoLR + isample + post-proc | 0.887 | 0.983 | 0.924 | 0.861 | 0.953 | 0.830 | 0.947 | 0.855 | 0.927 | 0.903 |
| Inter-annotator agreement | 0.859 | 0.973 | 0.917 | 0.823 | 0.965 | 0.673 | 0.934 | 0.810 | 0.877 | 0.857 |

Table 1: Dice scores for our proposed method on the hold out test set, and for other automatic multi-organ segmentation methods and inter-annotator agreement results [8] on the VISCERAL dataset online test set (now unavailable).

## 4 Conclusion

We proposed and evaluated a sampling scheme to deal with large images such as 3D CT scans. As shown in Section 3 the sampler enables fast training, and our results indicate that the final generalisation performance can be improved. To overcome manual hyperparameter tuning of the learning rate schedule we proposed an automated learning rate scheduler that is competetive with handcrafted schedules. Our experimental results suggests our combined algorithm gives new state-of-the-art performance for several organs, on the VISCERAL benchmark, and improves upon many human inter-annotator agreement scores.

**Appendix CNN setup:** For the dual path network architecture we build on several previous ideas [1, 9]. Compared to [1], we develop the architecture by replacing standard convolution layers with resnet blocks [9], and increase the maximum network depth from 11 to 21 layers. By having a deeper network and a down sampled pathway with input resolution 1/4 of the original resolution, we obtain a large receptive field of size $123^3$ whilst maintaining a deep high resolution pathway that does not compromise the resolution through pooling. This results in a total of 649,251 parameters. A sketch of the architecture is shown below where numbers inside round brackets give the input dimensions of each block. Numbers in square brackets refer to the number of feature maps used at each layer. The proposed configuration is memory efficient and allows for a large number of samples ($3D$ patches) per batch to ensure balanced class sampling and effective optimization, whilst maintaining a deep and wide enough network to capture the high variability and spatial semantics of the data.

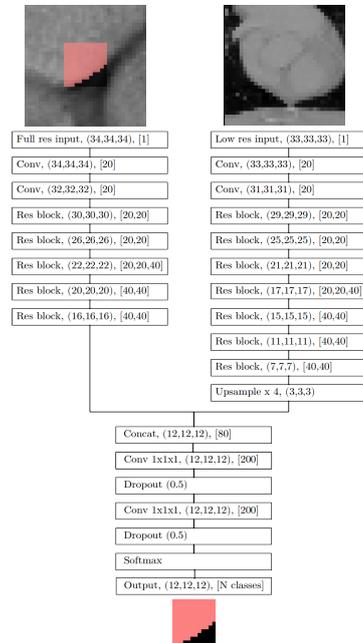

The blocks labeled 'Conv' are standard convolutional layers with kernel size $3 \times 3 \times 3$. The blocks labeled 'RA_block' and 'RB_block' are standard and bottleneck resnet blocks, respectively, as detailed in [9]. Each fully connected layer is preceded by a dropout layer with probability 0.5, and a softmax nonlinearity is used as a final classification layer. During training we perform data augmentation by re-sampling the 3D patches to a $[1mm, 1mm, 1.5mm] + \mathcal{U}(-0.1, 0.1)$ resolution. We also rotate each patch by $[\mathcal{U}(-10, 10), \mathcal{U}(-4, 4), \mathcal{U}(-4, 4)]$ degrees. We set voxels with values greater than 1000 to 1000, and values less than $-1000$ to $-1000$, and divide all values by a constant factor of 218 (the standard deviation of the dataset). We use Glorot initializations on all convolution layers and impose $L_2$ weight decay of size 0.0001, on all convolutional layers except on the last fully convolutional layer before the final softmax non-linearity. We use SGD with Nestrov momentum set at 0.8. We run each epoch for 100 batches and employ a linear learning rate warm up schedule, at the start of training and when changing learning rates, for a duration of 10 epochs. We use a standard cross-entropy loss function. For Algorithm 2 we chose to use the following configuration, $R_0 : \eta_0 = 0.05, \lambda_0 = 2$, $R_1 : \eta_1 = 0.01, \lambda_1 = 1$ and $R_2 : \eta_2 = 0.005, \lambda_2 = 0.5$. We found that having one run that explores at a higher learning rate $(R_0)$, one continuing at the same as the previously used learning rate $(R_1)$, and one that explores at half the learning rate $(R_2)$ worked well, and allowed the algorithm to effectively adapt to a well performing learning rate schedule.